\documentclass{article}

\usepackage{arxiv}

\usepackage[utf8]{inputenc} % allow utf-8 input
\usepackage[T1]{fontenc}    % use 8-bit T1 fonts
\usepackage{hyperref}       % hyperlinks
\usepackage{url}            % simple URL typesetting
\usepackage{booktabs}       % professional-quality tables
\usepackage{amsfonts}       % blackboard math symbols
\usepackage{nicefrac}       % compact symbols for 1/2, etc.
\usepackage{microtype}      % microtypography

\usepackage[authoryear]{natbib}
\usepackage{color}

\usepackage{amsmath}

\usepackage{listings}
\usepackage{subfigure}
\usepackage{graphicx}
\usepackage{caption}
\usepackage{pifont}

\definecolor{mygreen}{rgb}{0,0.6,0}
\definecolor{mygray}{rgb}{0.5,0.5,0.5}
\definecolor{mymauve}{rgb}{0.58,0,0.82}

\title{NGEMM: Optimizing GEMM for Deep Learning via Compiler-based Techniques}

\newcommand{\ignore}[1]{}

\newif\ifsubmit
\submittrue
\ifsubmit
    \newcommand{\liwen}[1]{}
    \newcommand{\yang}[1]{}
    \newcommand{\wenlei}[1]{}
    \newcommand{\ke}[1]{}
    \newcommand{\todo}[1]{}
\else
    \definecolor{gray}{rgb}{0.66, 0.66, 0.66}
    \definecolor{lightcyan}{rgb}{0.7, 0.7, 1.0}
    \newcommand{\liwen}[1]{[{\color{red}LWC: #1}]}
    \newcommand{\yang}[1]{[{\color{red}YC: #1}]}
    \newcommand{\wenlei}[1]{[{\color{red}WB: #1}]}
    \newcommand{\ke}[1]{[{\color{gray}EB: #1}]}
    \newcommand{\todo}[1]{[{\color{red}TODO: #1}]}
\fi

\usepackage{authblk}

\author[]{Wenlei Bao}
\author[]{Li-Wen Chang}
\author[]{Yang Chen}
\author[]{Ke Deng}
\author[]{Amit Agarwal}
\author[]{Emad Barsoum}
\author[]{Abe Taha}
\affil[]{Microsoft}
\affil[]{\texttt{\{Wenlei.Bao, LiWen.Chang, yanchen, kedeng, amitaga, ebarsoum, abetaha\}@microsoft.com}}

\begin{document}
\maketitle

\begin{abstract}
\ignore{Inference performance is important for deep neural network (DNN) services hosted in the cloud or edge, as these services often require quick response time.
High latency of DNN models may prevent them from being deployed to the production.}
\ignore{Moreover, even training suffers from high time cost for large and complex models. }
Quantization has emerged to be an effective way to significantly boost the performance of deep neural networks (DNNs) by utilizing low-bit computations. 
Despite having lower numerical precision, quantized DNNs are able to reduce both memory bandwidth and computation cycles with little losses of accuracy.
\ignore{The advantages demonstrate that neural networks with lower numerical precision is a promising way to pursue. }
\ignore{Moreover, hardware companies also enable the support for lower numerical precision training and inference in their processors and accelerators.}
Integer \textit{GEMM} (General Matrix Multiplication) is critical to running quantized DNN models efficiently, as GEMM operations often dominate the computations in these models.
\ignore{Therefore, improving integer GEMM can have significant impact to the overall performance of the DNN models.}
Various approaches have been developed by leveraging techniques such as vectorization and memory layout to improve the performance of integer GEMM.
However, these existing approaches are not fast enough in certain scenarios.
\ignore{\emph{GEMM} (General Matrix Multiplication) as one of the most compute intensive operations takes majority of the time spent in deep learning inference and training.} 
\ignore{Its performance has been optimized and provided in BLAS (Basic Linear Algebra Subprograms) libraries.}
\ignore{However, quantized GEMM operations with lower numerical precision are not well studied and supported efficiently.}

We developed \emph{NGEMM}, a compiler-based GEMM implementation for accelerating lower-precision training and inference.
NGEMM has better use of the vector units by avoiding unnecessary vector computation that is introduced during tree reduction. 
\ignore{NGEMM takes advantage of the instructions for low-precision computations supported by CPU.
but can be extended to other hardware and accelerator with similar instruction set provided.}
We compared NGEMM's performance with the state-of-art BLAS libraries such as MKL.
Our experimental results showed that NGEMM outperformed MKL \emph{non-pack} and \emph{pack} version by an average of 1.86x and 1.16x, respectively.
%Our experimental results showed that NGEMM outperformed the state-of-art %BLAS libraries such as MKL by an average of 1.4x.
%
We have applied NGEMM to a number of production services in Microsoft.

\end{abstract}

% keywords can be removed
\keywords{Deep Learning\and Training\and Inference\and Low-Precision\and GEMM\and Compiler}

\section{Introduction}

In the past few years, deep neural networks (DNNs) have demonstrated great success in various domains such as natural language processing~\citep{howard2018universal, radford2018improving}, speech recognition~\citep{deng2013new, anwar2015fixed} and computer vision~\citep{simonyan2014very, szegedy2015going}.
The increasing size and complexity of DNN models have become an impediment to the deployment of these models on edge devices where latency is often a hard requirement.
Even for cloud platforms like Azure, AWS and Google Cloud, where more computational resources are available, complex models such as BERT~\citep{devlin2018bert} may still cause long service time and high cost for both training and inference. 
\ignore{This trend may continue.}

\ignore{Various approaches have been proposed to improve the performance of DNN models~\citep{lin2016fixed, ioffe2015batch,han2015learning}. Among these approaches, quantization has demonstrate its success and become more popular. Quantization technique uses lower numerical precision for both deep learning training and inference, 8-bit for inference and 16-bit for training compared to 32-bit floating point precision, with marginal cost or no cost in accuracy~\citep{lin2015neural}.}

Quantization has emerged to be an effective way to significantly improve the performance of DNN models by utilizing low-bit computations~\citep{raghuraman2018quantization, jacob2018quantization, rastegari2016xnor, han2015learning, lin2016fixed}. 
By converting the weights and activations of a DNN model from high-precision floating point values to low-precision representations, quantization is able to reduce the model size and hence requires less memory for running the model.
Smaller memory footprint can have better cache behavior, because more intermediate computation results can be kept in the cache for re-use.
Moreover, computations on lower numerical representations such as 8-bit integers almost always need fewer clock cycles on contemporary general purpose CPUs and GPUs, compared with their high precision counterparts such as 32-bit floating point.
\ignore{Furthermore, there is little losses of accuracy in spite of the reduced numerical precision.}

\ignore{Despite the advantages of low-precision training and inference 
\ignore{quantizated models with lower numerical precision have better speed performance because of higher memory bandwidth and less numerical computations},however, it is not trivial to deploy the reduce-precision inference for several reasons~\citep{park2018deep}. 
When deep learning inference is related to core services in production, the accuracy loss has to be in control, which means the accuracy loss of reduce-precision inference compared to floating-point has to be within limited percentage range. In the meanwhile, the model parameters obtained from low-precision training also need to be determined carefully.}

GEMM operations often dominate the computation for training or inferencing DNN models~\citep{jia2014learning}.
Consequently, high performant integer GEMM has great impact to the efficiency of quantized DNN models, where high-precision floating point matrix elements are converted to low-bit such as 8-bit integers.
Recently, there have been increasing efforts in developing high-performance low-precision GEMM libraries, such as FBGEMM~\citep{fbgemm}, gemmlowp~\citep{jacob2017gemmlowp} and MKL~\citep{mkl}, where techniques such as vectorization, memory layout and various tiling schemes~\citep{goto2008anatomy, van2015blis} are applied to optimize GEMM computation.
However, they are still not fast enough in certain scenarios.
\ignore{However, these solutions are not optimal as they do not fully utilize the available processing units.}
 
\ignore{However, quantized GEMM operator with low numerical precision is not well studied in academia, and also not supported efficiently in current high-performance BLAS libraries. These libraries provides GEMM for general purpose but not special optimized for deep learning scenarios.
Particularly, they did not optimize shapes and sizes of matrices, which are common in deep learning inference and training, also they did not take advantage of the nature of weight matrix, which are constant during inference and highly reused in training.}

We developed NGEMM, an optimized GEMM implementation for DNN models based on compiler techniques.
NGEMM can provide high performance GEMM computation with different low-precision representations for various target machines.
Our experimental results showed that NGEMM outperformed MKL non-pack and pack version by an average of 1.86x and 1.16x respectively.
NGEMM has been used in a number of Microsoft production services~\citep{chang2018accelerating, onnxruntime}\footnote{NGEMM is used as external custom functions for specific integer BLAS routines in our Halide-based inference engine~\citep{chang2018accelerating}, and is implemented as a rewriting pass in code generation of Nuphar Execution Provider in ONNX Runtime~\citep{onnxruntime}.}.
 
In the rest of the paper, we will be focusing on Intel X86 CPUs, which are ubiquitous in modern computing systems. However, our methodology is widely applicable and not just specific to Intel CPUs.

\section{Background}

In this section, we briefly discuss the conventional approach to implementing low-precision GEMM.

\subsection{Conventional Approach}
%%%
\begin{figure*}[t]
\begin{minipage}{\textwidth}
\centering
\includegraphics[scale=0.7]{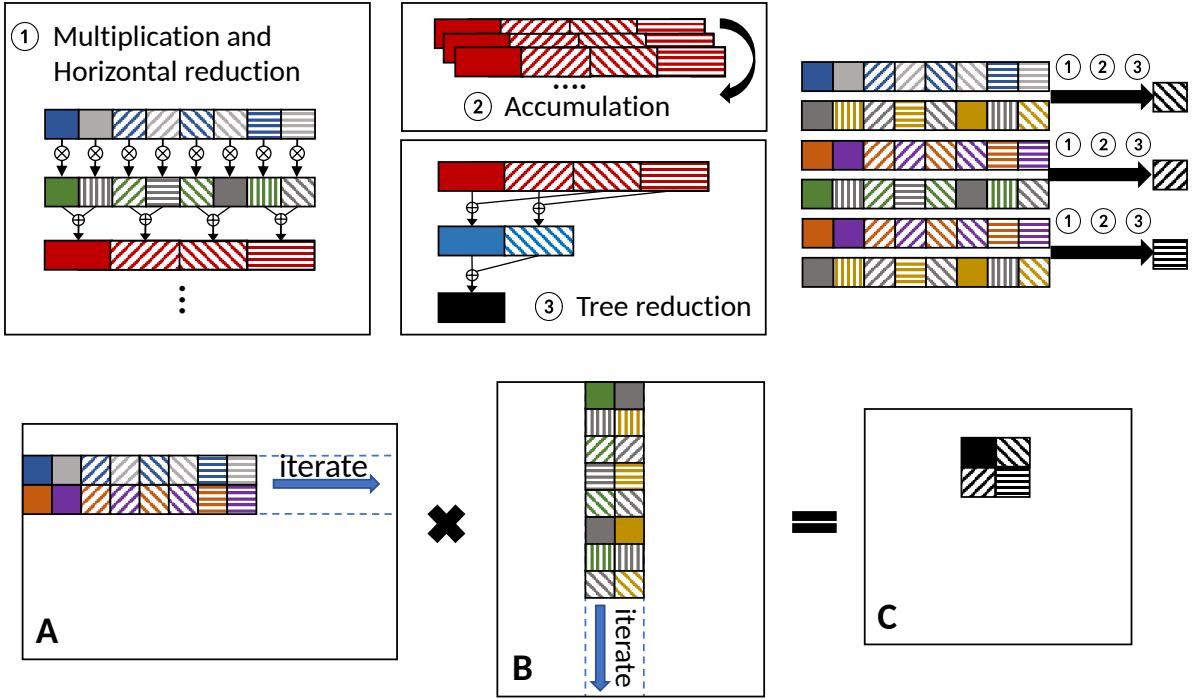}\vspace{0ex} \\
\end{minipage}%
\caption{Conventional GEMM}
\label{fig:gemm}
\vspace{-2ex}
\end{figure*}
%%%
Figure~\ref{fig:gemm} shows the conventional approach~\citep{rodriguez2018lower} to conduct low-precision GEMM.
Matrix A is the weight matrix with size of $[M \times K]$, and is typically determined from the training.
Matrix B is $[N \times K]$ input matrix, which contains the input data.
Matrix C $[M \times N]$ is the result. 
In low-precision GEMM, matrices A and B are the ones with low numerical precision such as 8- or 16-bit, while matrix C is in higher precision like 32-bit.
Throughout the example below, we will use unsigned 8-bit integers for matrix A and signed 8-bit integers for matrix B.
Other integer types have similar processes.

In Figure~\ref{fig:gemm}, \ding{192}, \ding{193} and \ding{194} are the typical phases to perform GEMM vector reduction, which computes the dot product of one row from matrix A and one column from matrix B to produce one element of matrix C.
Phase \ding{192} shows the process of converting low-precision values to higher precision. 
\texttt{VPMADDUBSW} takes two 8-bit integer vectors, $\mathbf{a_{u8}}=[a_0,a_1,...,a_{i-1}]$ and $\mathbf{b_{i8}}=[b_0,b_1,...,b_{i-1}]$, which contain unsigned and signed integers loaded from matrix A and B, respectively.
It then multiplies each unsigned 8-bit integer in $\mathbf{a_{u8}}$ and the corresponding signed 8-bit integer in $\mathbf{b_{i8}}$, produces a signed 16-bit integer.
Lastly, it adds adjacent pairs of the 16-bit integers to generate another vector $\mathbf{c_{i16}}$ :
\begin{align*}
\mathbf{c_{i16}}=[&c_0,c_1,...,c_{j-1}] \\  
=[&a_0*b_0+a_1*b_1,...,a_{i-2}*b_{i-2}+a_{i-1}*b_{i-1}]
\end{align*}

\ignore{Here $\mathbf{a_{u8}},\mathbf{b_{i8}}$ and $\mathbf{c_{i16}}$ have the same vector width in bits, but different number of elements $j=i/2$ because of the data type.}
\ignore{Since \texttt{VPMADDUBSW} requires one input to be of unsigned 8-bit integer and the other input to be signed 8-bit integer, the matrices A and B have to be in correct type to utilize the instruction.}

After that, \texttt{VPMADDWD} takes $\mathbf{c_{i16}}$ and a unit vector $[1,1,...,1]$ to perform horizontal reduction, which produces a signed 32-bit integer vector $\mathbf{e_{i32}}$ :
\begin{align*}
\mathbf{e_{i32}} = [&e_0, e_1, ... , e_{k-1}] \\
  = [&c_0+c_1, c_2+c_3, ... , c_{j-2}+c_{j-1}] \\
  = [&a_0*b_0+a_1*b_1+a_2*b_2+a_3*b_3, ... , \\ 
     &a_{i-4}*b_{i-4}+a_{i-3}*b_{i-3}+a_{i-2}*b_{i-2}+a_{i-1}*b_{i-1}]
\end{align*}

where each 32-bit value in $\mathbf{e}$ comes from four 8-bit values, thus $k = j/2 = i/4$.
The result of phase \ding{192} is a 32-bit integer vector $\mathbf{e_{i32}}$ with size of $k$.

Note that the vector width (i.e. the number of cubes in the vector) in Figure~\ref{fig:gemm} is just for illustration purpose.
The actual width depends on the data type and target machine.
For example, on an AVX-2 machine, where the vector width is 256 bits, the vector size $k$ of $\mathbf{e_{i32}}$ is $8$.
In contrast, for an AVX512 machine, $k$ becomes $16$. 

Phase~\ding{193} repeats the computation in Phase~\ding{192} multiple times for the rest of the bytes of the same row from matrix A and the same column from matrix B.
Each repetition produces a 32-bit vector $\mathbf{e_{i32}}$, which will be accumulated using instruction \texttt{VPADDD}.
This phase gives us: $\mathbf{e} = \sum_{K}{\mathbf{e_{i32}}}$.

%\begin{align*}
%    \bold{e_{sum}} = \sum_{iterate K}{\bold{e}}
%\end{align*}
%After the accumulation, the $\bold{e_{sum}}$ contains all

Phase~\ding{194} sums all of the 32-bit integers within the vector using tree-reduction to generate one final element in matrix C (shown as the black cube).
There are different ways to achieve this phase depends on instructions used. 
One approach is to use \texttt{VPHADDD}, which is the instruction to horizontally add adjacent pairs of 32-bit integers. Depending on vector width, multiple \texttt{VPHADDD} are needed to accumulate the integer values and generate the final value. 
The other approach is to utilize the \texttt{VPSHFD} instruction, which shuffles the 32-bit integers within the vector, with following \texttt{VPADDD} to accumulate corresponding values to generate the final result.

Finally, phase~\ding{192}~\ding{193}~\ding{194} are applied to all rows of matrix A and columns of matrix B to generate the whole matrix C.
Although these phases change the common reduction order of GEMM, they are typically applicable in integer computation unless overflow happens due to a very large $K$,
which is rare for most DNN models, where $K$ ranges from 100s to 1000s.\ignore{\yang{citation?}}

\ignore{Figure~\ref{fig:gemm} also shows similar pattern of phase \ding{192}~\ding{193}~\ding{194} to produce the other 3 elements to generate a $2\times 2$ cube in matrix C.}

\ignore{Besides the unsigned 8-bit and signed 8-bit integer example, signed 16-bit integer GEMM follow similar procedure but only need \texttt{VPMADDWD} in phase~\ding{192}, and Other low-precision GEMM with both unsigned 8-bit matrices also applies by casting unsigned 8-bit integers to signed 16-bit ones on-the-fly.}

\section{Methodology}

\subsection{Alternative Approach}

The conventional approach described in the previous section is straightforward to perform low-precision GEMM computation.
However, it is not optimal because the vector units are not fully used during the tree-reduction phase~\ding{194}.
Therefore, we present NGEMM, which has better use of the vector units by avoiding non-efficient vector computation such as tree-reduction.

%%%
\begin{figure*}[t]
\begin{minipage}{\textwidth}
\centering
\includegraphics[scale=0.7]{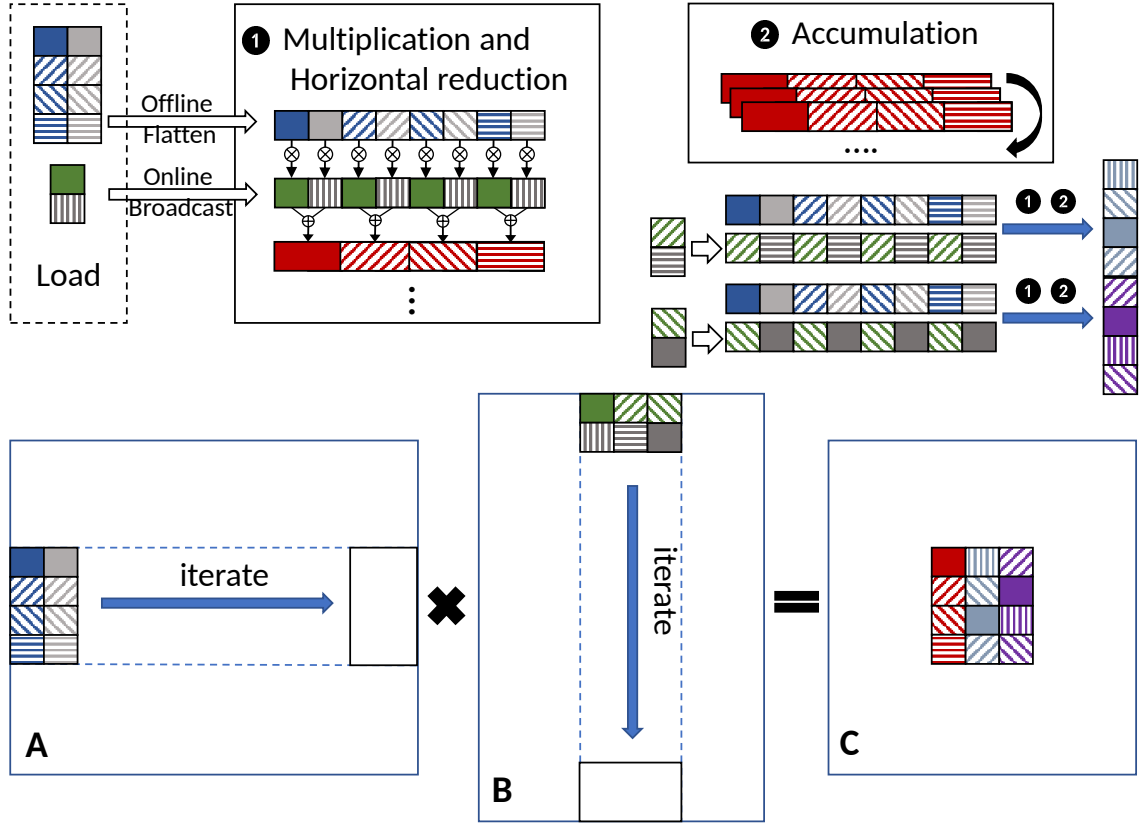}\vspace{0ex} \\
\end{minipage}%
\caption{NGEMM illustration}
\label{fig:ngemm}
\vspace{-2ex}
\end{figure*}
%%%

Figure~\ref{fig:ngemm} shows the details of our NGEMM.
We use the same configuration as that we used for the conventional approach: matrix A is the $[M \times K]$ weight matrix, matrix B is the $[N \times K]$ input matrix, and matrix C $[M \times N]$ is the result matrix.
Matrices A and B are in low numerical precision, and matrix C has 32-bit full precision. 

Similar to phase~\ding{192} in the conventional approach, phase \ding{182} computes partial results from two vectors, $\mathbf{a_{u8}}=[a_0,a_1,...,a_{i-1}]$ and $\mathbf{b_{i8}}=[b_0,b_1,...,b_{p-1}]$, using \texttt{VPMADDUBSW} and \texttt{VPMADDWD}. 
However, the shapes of the loaded data are different from the conventional approach even with the same vector length.
\ignore{$\mathbf{a_{u8}}$ can contain the data from multiple rows and columns of matrix A. }
In this illustration, its shape is $[4 \times 2]$, which is determined by the vector width and the width of horizontal reduction in practise. 
For example, the shape is $[8 \times 4]$ for AVX2.
Because both dimensions are shorter than the vector width, loads of $\mathbf{a_{u8}}$ access non-contiguous memory, which may compromise memory bandwidth.
To resolve the issue, we apply data layout transformation (see Sec~\ref{sec:layout}) to force $\mathbf{a_{u8}}$ to be fully contiguous in memory.

Furthermore, because vector $\mathbf{b_{i8}}$ has shorter length than $\mathbf{a_{u8}}$ ($p<i$),
we broadcast $\mathbf{b_{i8}}$ to $\mathbf{b'_{i8}}$: 
\begin{align*}
    \mathbf{b'_{i8}}= [b_0,b_1,...b_{p-1},...,b_0,b_1,...,b_{p-1}] 
\end{align*}
where $\mathbf{b'_{i8}}$ has the same width as $\mathbf{a_{u8}}$.
For example, if $p=4$, then $\mathbf{b'_{i8}}= [b_0,b_1,b_2,b_3,...,b_0,b_1,b_2,b_3]$, 
\ignore{The reason of the broadcasting is the different layout of vector $\mathbf{a_{i8}}$, which will reuse $\mathbf{b_{i8}}$.} 
Consequently, $\mathbf{c'_{i16}}$ and $\mathbf{e'_{i32}}$ will be as follows:
\begin{align*}
    \mathbf{c'_{i16}} = [&c_0,c_1,c_2,...,c_{j-1}] \\
    =[&a_0*b_0+a_1*b_1,a_2*b_2+a_3*b_3,...,\\ &a_{i-4}*b_{0}+a_{i-3}*b_{1},a_{i-2}*b_{2}+a_{i-1}*b_{3}]\\
    \mathbf{e'_{i32}} = [&e_0,e_1,e_2,...,e_{k-1}] \\
    =[&a_0*b_0+a_1*b_1+a_2*b_2+a_3*b_3,..., \\
    &a_{i-4}*b_0+a_{i-3}*b_1+a_{i-2}*b_2+a_{i-1}*b_3]
\end{align*}

Figure~\ref{fig:ngemm} shows an example of $p=2$ (two elements during load of matrix B) for simplicity, which is the 16-bit scenario:
\begin{align*}
   \mathbf{a'_{i16}} = [&a_0,a_1,...,a_{i-1},a_i] \\
   \mathbf{b'_{i16}} = [&b_0,b_1,...,b_0,b_1] \\
   \mathbf{c'_{i32}} = [&c_0,c_1,c_2,...,c_{j-1}] \\
    =[&a_0*b_0+a_1*b_1,a_2*b_0+a_3*b_1,...,\\ &a_{i-4}*b_{0}+a_{i-3}*b_{1},a_{i-2}*b_{0}+a_{i-1}*b_{1}]
\end{align*}
where only \texttt{VPMADDWD} is required in phase~\ding{182} to accumulate to 32-bit.

The second phase~\ding{183} accumulates all the partial results of $\mathbf{e'_{i32}}$ to directly form the final result vector in matrix C.

Figure~\ref{fig:ngemm} also shows the two more partial results obtained by applying phases~\ding{182} and~\ding{183} to the same part of A but different two vectors of B.
The same computation will be applied to the entire A and B and obtain the final result matrix C.

Compared with phases~\ding{192}~\ding{193}~\ding{194} of the conventional approach, our approach only has phases~\ding{182} and~\ding{183} but missing the tree-reduction phase, which is implicitly finished through the vector additions in phase~\ding{183}.
\ignore{This is because the process of tree-reduction is to calculate the summation of the vector to generate a single 32-bit integer, which is achieved by phase \ding{183} in the way of vector addition instead, and thus have a better vector efficiency.}
\ignore{We will have a separate comparison analysis of the efficiency in following section.}

Our presented approach is not limited to GEMM operations, but benefits other similar computations such as GEMV. 
Moreover, it is also applicable to GEMM operations with other types such as signed 16-bit or unsigned 8-bit integers.

It is worthwhile to mention that Intel adopted similar approach in MKL~\citep{intelblog},
which is contemporary work with our NGEMM.

\subsubsection{Data Layout}
\label{sec:layout}

We describe the data layout used by our approach in detail.
The layout is mainly for the weight matrix in a DNN model, e.g. matrix A in our examples.
Therefore, data marshalling can often be performed offline.
For those uncommon cases where both matrices are inputs, we can simply perform online packing with extra overhead.

\ignore{For input matrix, its layout is the same with the conventional approach, but slightly different in computation, which is described above.} 

%%%
\begin{figure*}[t]
\begin{minipage}{\textwidth}
\centering
\includegraphics[scale=0.8]{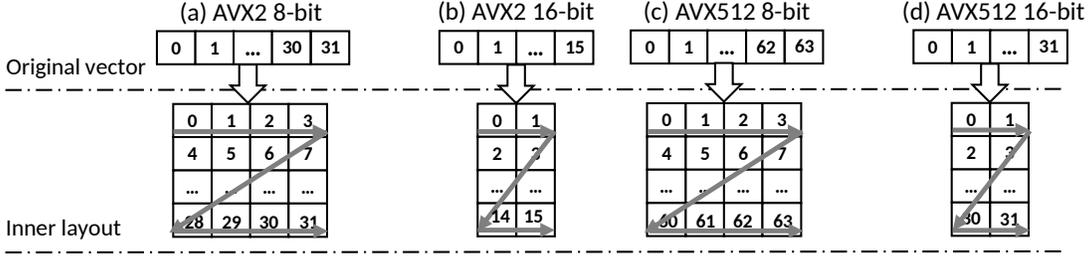}\vspace{0ex} \\
\end{minipage}%
\caption{NGEMM inner layout for weight matrix}
\label{fig:layout}
\vspace{-2ex}
\end{figure*}
\begin{figure*}[t]
\begin{minipage}{0.5\textwidth}
\centering
\includegraphics[scale=0.8]{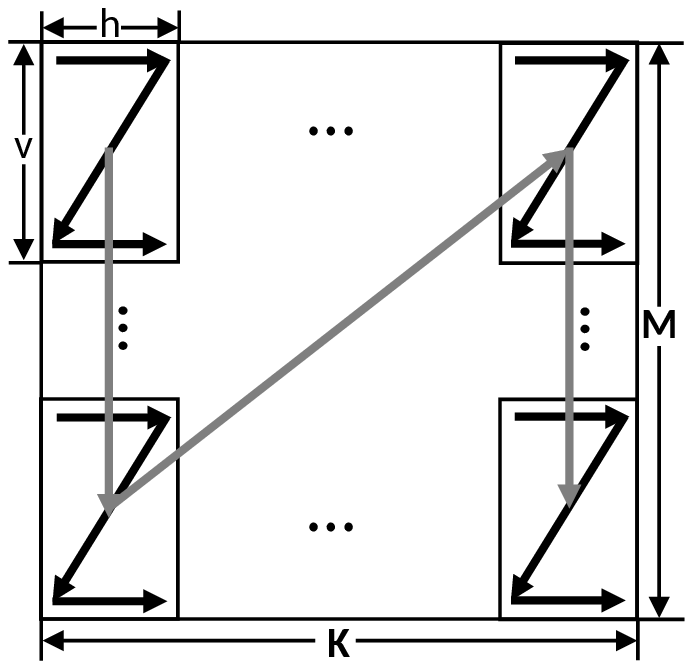}\vspace{0ex} \\
(a) Outer layout without tiling
\end{minipage}%
\begin{minipage}{0.5\textwidth}
\centering
\includegraphics[scale=0.8]{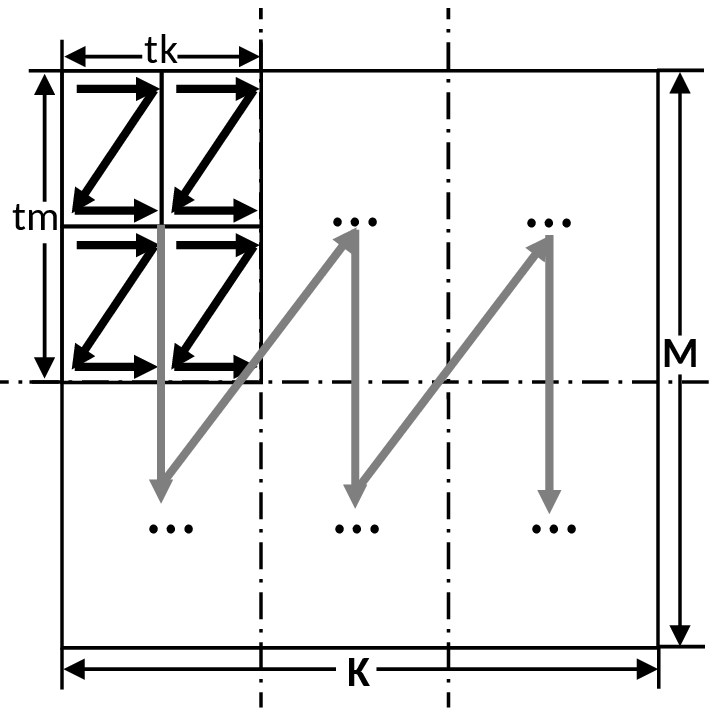}\vspace{0ex} \\
(b) Outer layout with tiling
\end{minipage}%
\caption{NGEMM outer layout for weight matrix}
\label{fig:tiling}
\vspace{-2ex}
\end{figure*}

%%%
Figure~\ref{fig:layout} and Figure~\ref{fig:tiling} demonstrates the two-level hierarchy of the layout used in NGEMM.
Figure~\ref{fig:layout} shows various inner layouts in NGEMM, where the top row display the original vector access.
Assuming the memory address grows horizontally, the original vector loads the contiguous elements from the memory.
Vector width $w$ depends on hardware instructions and data types as displayed in Figure~\ref{fig:layout}:
\begin{align*}
  w = \frac{Instruction Width In Bits}{Data Type In Bits} 
  =\left\{
  \begin{array}{r@{}l}
        InstWidth_{\text{AVX2}}&/Integer_{8bit} = 32 \\
        InstWidth_{\text{AVX2}}&/Integer_{16bit} = 16 \\
        InstWidth_{\text{AVX512}}&/Integer_{8bit} = 64 \\
        InstWidth_{\text{AVX512}}&/Integer_{16bit} = 32
  \end{array}
  \right.
\end{align*}

The bottom row draws the inner layout, which determines the data access pattern for the matrix. 
Instead of accessing data from a single row or column, inner layout will first access $h$ elements along $K$ dimension and then move to next row, and repeat this process $v$ times along M dimension.
This kind of memory access forms a small zigzag pattern, which is known as \textit{Morton code}~\citep{morton1966computer}.
Usually, $h$ and $v$ are determined by hardware instructions and data type:
\begin{align*}
  h= \left\{
  \begin{array}{ll}
    &4 \hspace{10pt} when \hspace{10pt} 8\text{bit integer} \\
    &2 \hspace{10pt} when \hspace{10pt} 16\text{bit integer}
  \end{array}
  \right.
  v= \frac{w}{h} \left\{
  \begin{array}{ll}
       &  8 \hspace{14pt} when \hspace{10pt} \text{AVX2}\\
       &  16 \hspace{10pt} when \hspace{10pt} \text{AVX512}
  \end{array}
  \right.
\end{align*}

The size of $h$ simply implies the number of elements needed to be accumulated to a single 32-bit value in phase (\ding{182}). 
\ignore{If $h < h_{min}$, it indicates that more instructions such as VPMADDUBSW and VPMADDWD are required in the up-convert process ( \ding{182} and \ding{192}).
If $h > h_{min}$, it indicates that more than one 32-bit value exist in inner layout vector along 
$K$ dimension, thus the \textit{tree-reduction} (phase \ding{194}) are required to accumulate these 32-bit values into one. 
}

Figure~\ref{fig:tiling} illustrates the outer layout pattern in NGEMM for weight matrix.
The outer layout can be combined with loop optimizations such as loop tiling to deliver better performance. 
Loop tiling is a classic loop transformation technique to maximize parallelism and improve cache locality~\citep{hong2016effective,bao2017analytical,bao2017efficient}. 
It has been widely adopted in various BLAS libraries to optimize intensive computation such as GEMM.

Figure~\ref{fig:tiling} (a) shows the outer layout without loop tiling applied, which iterate the inner layout pattern first through M dimension and then K dimension for the whole matrix.
\ignore{
There are two iteration orders: along M dimension (shown on the left) and along K dimension (shown on the right)
These two orders are aligned with loop permutation and will be discussed in the next section.
}
It is worth mentioning that if columns or rows are not multiple of $h$ or $v$, we simply pad 0-element, respectively.

%%%
%%%

Figure~\ref{fig:tiling} (b) shows the tiling with a tile size $tk$-by-$tm$ combined with the layout for matrix A, where $<tm,tn,tk>$ denote tiling sizes along $<M,N,K>$ dimension of the GEMM. 
In our tiling scheme, $tk$-by-$tm$ typically contains multiple $h$-by-$v$.
\ignore{
A typically relation between layout size $h$-by-$v$ and tiling size $tk$-by-$tm$ should have following constrains:
\begin{align*}
tk &= \alpha \times h \hspace{6pt} where \hspace{6pt} \alpha = 1,2,3,..., K/h\\
tm &= \beta \times v \hspace{6pt} where \hspace{6pt} \beta = 1,2,3,...,M/v
\end{align*}}
Thus the tiling size selection of $<tm,tn,tk>$ becomes $<\beta v,tn,\alpha h>$, where $tn$ is the tiling size along dimension N.
The outer layout then is to iterate the tile pattern through the whole matrix.

Moreover, we also apply inter- and intra- tile level optimizations. 
For inter-tile level, we perform loop unrolling to increase locality and reduce loop control overhead within a tile. 
For intra-tile level, different loop permutation can be used, which corresponds to the outer layout iterate order described previously.
Overall, the loop tiling size and permutation are a typical search-space-exploration problem, which can be solved by techniques such as auto-tuning.

\subsection{Latency Analysis}

In this part, we perform the latency analysis between the conventional approach and our NGEMM. 
Note, here, we only considers the computation cost, but ignores the cost of loads/stores, since after data layout transformation, memory accesses in the both methods have become contiguous.

Assume the matrix sizes are the same as previous section, i.e. $C_{M\times N}=A_{M\times K}\times B_{K\times N}$ with vector width $w$.
The latency of the conventional approach can be calculated as follows:

\begin{align*}
Latency_{convention} &= M \times (\frac{K}{w} \times (latency(\text{\ding{192}})+latency(\text{\ding{193}})) + latency(\text{\ding{194}})) \\
&= M \times \frac{K}{w} \times (latency(\text{\ding{192}})+latency(\text{\ding{193}})) + M\times (c_1 log(w))
\end{align*}

where $latency(\text{\ding{192}})$ and $latency(\text{\ding{193}})$ are the total cost for phase \ding{192} and \ding{193} in the conventional approach respectively.
Here, $c_1 log(w)$ is the latency cost of tree-reduction computation $latency(\text{\ding{194}})$, and $c_1$ is constant parameter.

For our NGEMM with layout size $h\times v = w$, the latency is as follows:

\begin{align*}
Latency_{\text{NGEMM}} &= \frac{M}{v} \times \frac{K}{h} \times (latency(\text{\ding{182}})+latency(\text{\ding{183}})) \\
&= M \times \frac{K}{w} \times (latency(\text{\ding{182}})+latency(\text{\ding{183}}))
\end{align*}

where latency(\text{\ding{182}}) and latency(\text{\ding{183}}) are the time cost of phase \ding{182} and \ding{183} in NGEMM. 
The speedup ratio between these two can be expressed as:

\begin{align}
 ratio_{speedup} = \frac{Latency_{convention}}{Latency_{\text{NGEMM}}}
 = \frac{K/w\times (latency(\text{\ding{192}})+latency(\text{\ding{193}})) + (c_1 log(w))}{K/w\times (latency(\text{\ding{182}})+latency(\text{\ding{183}}))}
\end{align}

where phases of \ding{192}~\ding{193} and \ding{182}~\ding{183} have the same latency cost, and their latency can be noted as $c_2$.
Then equation (1) can be expressed as:

\begin{align}
 ratio_{speedup} 
 = 1 + \frac{c_1 log(w)}{K/w\times c_2} = 1 + c_3 \cdot \frac{log(w)}{l}
\end{align}

where $l=K/w$ represents the reduction dimension in the step of vector width of matrix multiplication, $w$ is the vector width and $c_3=c_1/c_2$ is constant parameter.

From equation (2), we can observe that the speedup is relevant to reduction dimension  and vector width. 
The longer vector width and the shorter reduction dimension will result in better speedup.
It is worth mentioning that the trend of increasing vector width $w$ continues in new generation of processors. 

\section{Implementation}
\ignore{Implementation details of NGEMM are described in this section. }
\ignore{Our NGEMM is implemented under~\citep{onnxruntime}, which take the advantage of ONNX IR~\citep{onnx}.}

We implemented single-threaded NGEMM by leveraging TVM~\citep{chen2018tvm} tensorize schedule primitive that replaces partial computation code with the corresponding intrinsics in LLVM~\citep{lattner2004llvm}.
It can be easily extended to support \textbf{multi-threading} by parallelizing the outermost loop.
\ignore{This is a typical compiler approach to automatically generate the micro kernels rather than manually written them.}
\ignore{However, TVM tensorize is not quite technically mature. 
For example, it cannot handle tail condition properly, which is quite common scenarios in GEMM and necessary features in loop transformations~\citep{bao2016polycheck,bao2018compiler}. 
Besides, TVM has limited support of LLVM intrinsic, which miss many frequently used ones.}
% Yang: If not necessary, don't say some negative thing for other's work :) Let's skip this part and simply say what we did.
%
We modified TVM code to properly handle tail condition during code generation, which is important for certain loop transformation techniques~\citep{bao2016polycheck,bao2018compiler}.
Furthermore, we added several LLVM intrinsics that were not supported by TVM. 
\ignore{\yang{please double-check to make sure my description is accurate.}}
\ignore{Therefore, we developed NGEMM by editing TVM code base to add those features that are required to realize the implementation.}
It is worthwhile to mention that our NGEMM is not limited to TVM.
The same techniques can be implemented in other compiler-based frameworks ~\citep{rotem2018glow, mlir}. 

\textbf{Layout} The layout of the weight matrix has been prepacked offline, and the marshalled matrix is fed as a constant initializer to the inference runtime.
NGEMM generates different layouts for the weight matrix based on numerical precision requirements and supported instruction sets.
\ignore{Besides, the inner layout get flattened in memory in order to have better memory performance}

\textbf{Loop Optimization} 
The optimal loop tiling sizes and permutation orders are determined through compiler auto-tuning.
NGEMM generates the best choice based on matrix sizes, target machines and numerical precision.

Moreover, it is straightforward to explore more extensions such as \textbf{operation fusion} with the help of compiler techniques. 
DNN models have pre- and post-GEMM operations, which can be fused together with GEMM to increase cache locality and thus improve performance. 
\ignore{FBGEMM~\citep{fbgemm} provides similar fusion features.}

\ignore{
We can also apply more sophisticated \textbf{thread-binding} and \textbf{matrix-sharding} strategies. 
}
\ignore{We leave this as future work.}
\ignore{
\textbf{Parallelism} is another extension that could make direct impact of performance. Thread-level parallelism can be achieved directly by generating parallel loops for the outer most iteration of GEMM.
}

\ignore{
\yang{I commented out the entire loop optimization part because seems it has nothing to do with our implementation.}}

\section{Evaluation}
We evaluated NGEMM's performance against MKLML in MKL-DNN v0.21\footnote{This version of MKL fixed certain performance issues in previous versions.}.
Our NGEMM was implemented in ONNX Runtime~\citep{onnxruntime}, using a custom version of TVM with LLVM 9.0.0. 
The experiments were conducted on a machine with 8-core 2.3GHz Intel Xeon E5-2673V4 processors where AVX2 is supported.
The machine runs Ubuntu 16.04 and GCC 6.5.
No frequency scaling (DVFS) related techniques were used in the experiments~\citep{bao2016static,farkas2000quantifying}.
The performance numbers were obtained by taking ggeometric means across repetitions after several warmup runs. 
The numbers include only function calls of MKL rountines or our generated functions, without any ML framework overhead\footnote{Speedup numbers reported in the previous version include ONNX Runtime overhead.}.

%%%
\begin{figure*}[t]
\begin{minipage}{\textwidth}
\centering
\includegraphics[scale=0.7]{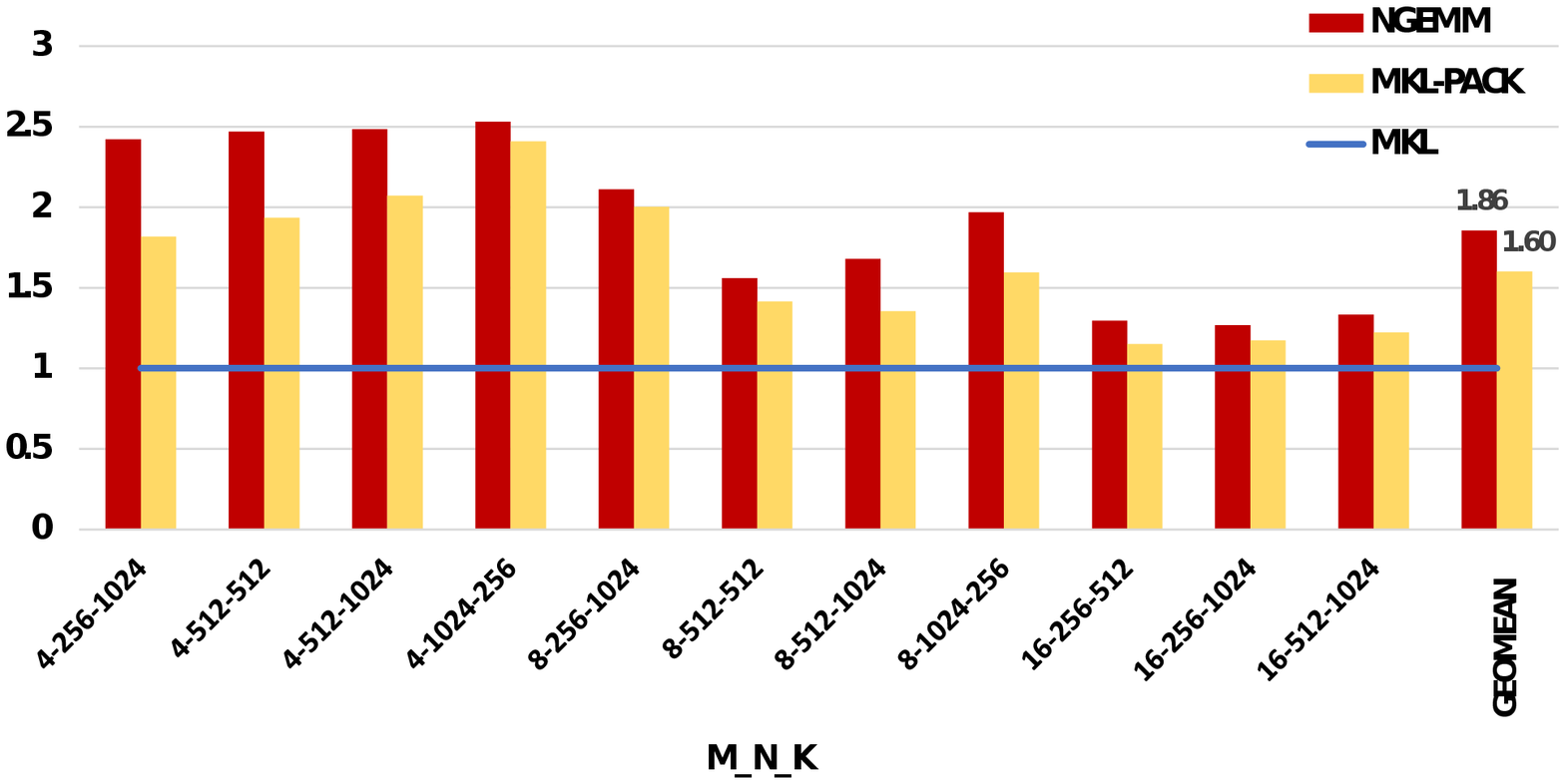}\vspace{0ex} \\
\end{minipage}%
\caption{Performance Speedups of NGEMM and MKL-PACK(single-threaded \texttt{cblas\_gemm\_s8u8s32\_compute} with a offline process \texttt{cblas\_gemm\_s8u8s32\_pack}) vs MKL(single-threaded \texttt{cblas\_gemm\_s8u8s32}) over different sizes}
\label{fig:results}
\vspace{-2ex}
\end{figure*}

\ignore{
\begin{minipage}{0.5\textwidth}
\centering
\includegraphics[scale=0.6]{fig/result2.png}\vspace{0ex} \\
(b) Production Models
\end{minipage}%
}
%%%

Figure~\ref{fig:results} shows the speedups of NGEMM and MKL-PACK (the pack version with the routines \texttt{cblas\_gemm\_s8u8s32\_compute} and the offline packing routines \texttt{cblas\_gemm\_s8u8s32\_pack}) over non-pack MKL (the version with the routines \texttt{cblas\_gemm\_s8u8s32}) on various problem sizes with 8-bit integers. 
Similar to NGEMM, MKL-PACK applies packing to the weight matrix.
In contrast, MKL does not do any packing.
\ignore {The bars in the figure are the speedups of NGEMM, while the line shows the MKL, where the speedup is always 1.0 compared to itself. }
The last two bars in the figure are the geometric means of the speedups of NGEMM and MKL-PACK over MKL across all the problem sizes.
NGEMM and MKL-PACK demonstrated 1.86x and 1.60x speedups compared to MKL, respectively.

As shown in the graph, NGEMM has better performance compared to MKL for all the problem sizes that we chosen. 
NGEMM also shows about 1.16X speedups over MKL-PACK.
Moreover, we observed a huge memory cost (about 12MB) with MKL-PACK for packing a [$512\times 512$] \texttt{int8} weight matrix, whereas NGEMM consumed only 0.25MB, which is the size of the matrix.

%But the speedup varies, small batch sizes have larger benefits, where the batch size is the matrix M dimension. During the evaluation, we find that even for large batch size, NGEMM still shows around 1.1X speedups compared to MKL.

It is worth mentioning that the actual performance benefits of NGEMM might vary for different deep learning models. 
Real models have computations performed by other ops such as element-wise ops, which consume considerable time. 
In our experiments with real production models, NGEMM could delivered up to 1.34X speedups over MKL without packing.

\ignore{
In Figure~\ref{fig:results} (b) presents the result of real production models, a 4-layered custom LSTMs (\texttt{Speech1}), a 6-layered custom LSTMs (\texttt{Speech2}), and a 6-layered bidirectional LSTMs (\texttt{Speech3}). 
All models had the sequence length of 40 and the batch size of 1.
}

\section{Related work}

Many research studies~\citep{vanhoucke2011improving, Rastegari_2016, mellempudi2017mixed, das2018mixed} have shown that low numerical precision can be applied to DNN models with minimal to no accuracy losses.
Quantization techniques have been adopted by many DNN frameworks to improve training and inference performance.
XLA (Accelerated linear algebra)~\citep{leary2017xla} is a compiler backend for TensorFlow~\citep{abadi2016tensorflow}, which supports various optimizations including quantization and generate machine instructions for different targets.
Glow~\citep{rotem2018glow} is a machine learning compiler, which consumes neural network graphs, performs high-level graph optimizations and low-level machine-dependent optimizations for diverse hardware targets. 
Glow also supports quantization with various bit-widths. 
TVM~\citep{chen2018tvm}, another compiler stack for deep learning, compiles the model into low-level \ignore{Halide-based} IR and performs loop optimizations. 
It supports multiple hardware backends and quantization with different bit-widths. 
Our work, NGEMM, incorporates with Microsoft ONNX Runtime~\citep{onnxruntime}.

% some example of previous work, reference from intel 

\ignore{
%\subsection{Relation to Deep learning frameworks}
With the benefits demonstrated by researchers across various models, many deep learning frameworks adopt the quantization techniques to improve their training and inference performance.

XLA (Accelerated linear algebra)~\citep{leary2017xla} is a compiler backend for TensorFlow~\citep{abadi2016tensorflow}, which support various optimizations including quantization and generate machine instructions for different targets.
Glow~\citep{rotem2018glow} is a machine learning compiler, which consume neural network graph, perform high level graph optimizations and low-level machine-dependent optimizations for diverse hardware targets. Glow also supports quantization with various bit-widths. 
TVM~\citep{chen2018tvm}, as a deep learning compiler stack, compiles the model into low-level Halide-based IR and perform loop optimizations. It supports multiple hardware backends and also different bit-width quantization.
}

Many \textbf{BLAS libraries}, meanwhile, have implemented low precision GEMM for various architectures. 
Intel MKL (Math Kernel Library)~\citep{wang2014intel} provides well tuned 8- and 16-bit GEMM kernels, which is widely adopted across many DNN frameworks as a CPU vendor library.
NVIDIA cuBLAS library~\citep{nvidia2008cublas} provides GPU counterparts on NVIDIA GPUs. 
FBGEMM~\citep{fbgemm} is a reduced-precision linear algebra library for deep learning inference, and is integrated into Caffe2. 
Gemmlowp~\citep{jacob2017gemmlowp} is a library that only supports low-precision GEMM.
In addition to execution time, it is also optimized to minimize the power consumption, which is crucial for edge devices.

Other than TVM, which is used to implement our techniques in the paper, general-purpose code generation frameworks~\citep{chill, spiral, tangram, changThesis17, cgo_kernel}, and domain-specific languages~\citep{halide, delite}, are also capable to adopt our techniques for optimizing GEMM routines. 

\ignore{
provides series optimized math functions including BLAS with 32-bit floating point precision to fully utilize Intel CPU features.
Nvidia cuBLAS library~\citep{nvidia2008cublas} supported BLAS functionality on NVIDIA GPU. It implements on top of the NVIDIA CUDA runtime with 32-bit full precision.

Besides, there are several libraries that support low numerical precision BLAS operations for quantization.
MKL-ML also supports well tuned 8-bit and 16-bit GEMM kernels, which are widely adopted across many deep learning frameworks as vendor libraries after quantization. 
FBGEMM~\citep{fbgemm} is a reduced-precision linear algebra library for deep learning inference. It is open-sourced by Facebook and integrated with Caffe2 deep learning framework.
Gemmlowp~\citep{jacob2017gemmlowp} is a library that only supported low-precision GEMM, but besides execution time, it also optimized to minimize the power consumption, which is crucial for edge devices.}

\section{Conclusion}

In this paper, we present NGEMM, an optimized low-precision GEMM implementation based on compiler techniques.
%
%NGEMM is particularly useful for DNN inference on CPUs. 
%
Compared to the conventional approach adopted by contemporary BLAS libraries, our approach does not require tree reduction and hence has better performance.
We implemented a hierarchical layout for the weight matrix to further improve the latency.
Our evaluation on various problem sizes demonstrate an average of 1.16X and 1.86X speedup over state-of-the-art MKL library with and without packing. A number of production models also show up to 1.34X speedup using NGEMM compared to MKL without packing.

\bibliographystyle{authordate1}  

\bibliography{ref}
\end{document}